\pdfoutput=1 % FOR ARXIV
\def\year{2021}\relax

\documentclass[letterpaper]{article} % DO NOT CHANGE THIS
\usepackage{aaai21}  % DO NOT CHANGE THIS
\usepackage{times}  % DO NOT CHANGE THIS
\usepackage{helvet} % DO NOT CHANGE THIS
\usepackage{courier}  % DO NOT CHANGE THIS
\usepackage[hyphens]{url}  % DO NOT CHANGE THIS
\usepackage{graphicx} % DO NOT CHANGE THIS
\usepackage{graphicx}  % DO NOT CHANGE THIS
\usepackage{natbib}  % DO NOT CHANGE THIS AND DO NOT ADD ANY OPTIONS TO IT
\usepackage{caption} % DO NOT CHANGE THIS AND DO NOT ADD ANY OPTIONS TO IT
\usepackage{bbm}
\usepackage{amsmath}

\title{Content Masked Loss: Human-Like Brush Stroke Planning in a Reinforcement Learning Painting Agent}
\author{

    Peter Schaldenbrand,
    Jean Oh
    \\
}
\affiliations{
    Carnegie Mellon University\\
    5000 Forbes Avenue\\
    Pittsburgh, PA 15213\\
    pschalde@andrew.cmu.edu,
    jeanoh@nrec.ri.cmu.edu

}

\begin{document}

% \linenumbers % Take out for camera ready

\maketitle

\begin{abstract}
The objective of most Reinforcement Learning painting agents is to minimize the loss between a target image and the paint canvas.  Human painter artistry emphasizes important features of the target image rather than simply reproducing it \cite{dipaola2007-painterly}.  Using adversarial or $L_2$ losses in the RL painting models, although its final output is generally a work of finesse, produces a stroke sequence that is vastly different from that which a human would produce since the model does not have knowledge about the abstract features in the target image.  In order to increase the human-like planning of the model without the use of expensive human data, we introduce a new loss function for use with the model's reward function: Content Masked Loss. In the context of robot painting, Content Masked Loss employs an object detection model to extract features which are used to assign higher weight to regions of the canvas that a human would find important for recognizing content. The results, based on 332 human evaluators, show that the digital paintings produced by our Content Masked model show detectable subject matter earlier in the stroke sequence than existing methods without compromising on the quality of the final painting. Our code is available at \url{https://github.com/pschaldenbrand/ContentMaskedLoss}.
\end{abstract}

\section{Introduction}

While there are numerous differences between the methods used by a skilled human painter and a machine printer, the end products may look similar.  As is the case with photo-realistic portraits, a human painter hopes to recreate a target image as closely as possible with paint.  A printer, though not using paint and a brush, would have a similar objective, however, would go about it in a completely different fashion.  A printer may use a simple heuristic to render the image such as starting from the top left and working left-to-right and top-top-bottom, but a human painter would likely take an abstract approach. As a painter paints using the commonly used ``blocking in" \cite{Scott2017-blockingIn} technique, you can recognize the subject of the painting early in the painting process, and the fidelity of the rendering improves with time.  In this way, painting can be considered an anytime algorithm: If the painting process is interrupted at any point, a valid solution should be available, but the solution becomes better the longer the process executes for.

The goal of Stroke Based Rendering (SBR) is to find methods to reproduce a target image using visual elements, such as paint brush strokes.  Many SBR methods \cite{Hertzmann2003-strokeBasedRenderingSurvey} were deterministic algorithms which placed strokes onto a canvas that would minimize the loss between the next canvas and the target image. Modern SBR methods \cite{Ganin2018-spiral,Huang2019-learningToPaint,Nakano2019-neuralPainters,Xie2012-artistAgent} use Reinforcement Learning models to generate the stroke sequences.  All of these RL methods use an adversarial or $L_2$ loss in their reward functions which are designed to minimize the difference between the canvas and the target image.  These loss functions are oblivious to abstract content of the image and the stroke sequence can follow any method that the model finds on its way to minimizing the loss.

\begin{figure}[ht]
    \centering
    \includegraphics[width=\columnwidth]{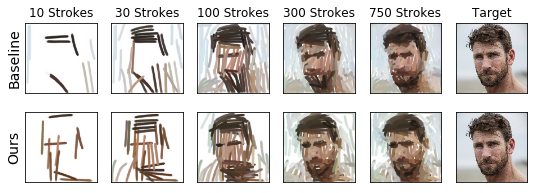}
    \caption{Our model was trained using a clipped $L_1$ loss with Content Masking in its reward function, while the baseline used adversarial loss.}
    \label{fig:ours_vs_baseline}
\end{figure}

\begin{figure}
    \centering
    \includegraphics[width=\columnwidth]{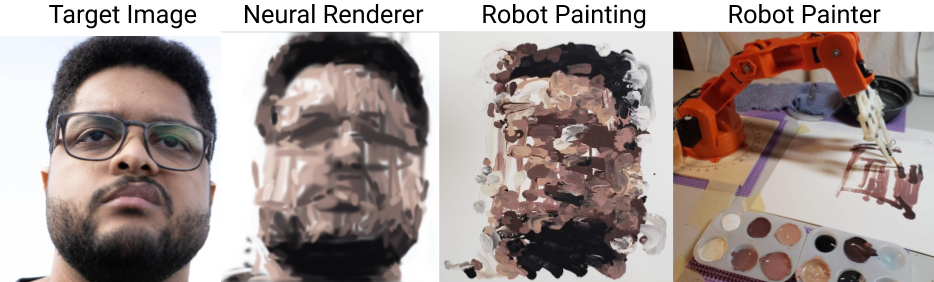}
    \caption{The Reinforcement Learning Painter generated brush strokes to paint the target image.  The strokes can be renderered into an image by feeding them to the Neural Renderer or the robot painting agent as seen above. The robot painter is a TinkerKit Braccio robotic arm.  10 paint colors, that are chosen by running k-means clustering on the target image, are mixed by a human and provided to the robot.  The arm has a single paint brush and can clean it using water and a towel. The robot is fully autonomous while painting.}
    \label{fig:robot_painting_example}
\end{figure}

\begin{figure}[ht]
    \centering
    \includegraphics[width=\columnwidth]{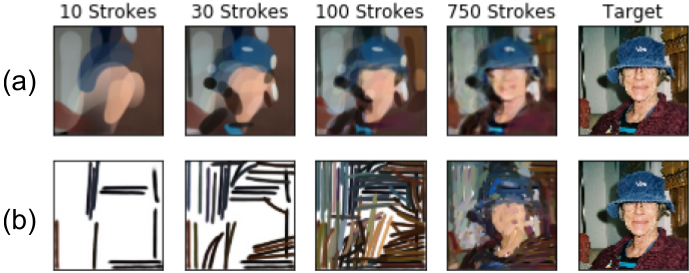}
    \caption{Row (a) is the \cite{Huang2019-learningToPaint} model and row (b) is our baseline model trained using a constrained Neural Renderer. The constraints assume that the brush strokes are fixed width, opaque, and do not exceed a maximum length.  Our baseline model also assumes a white starting canvas whereas \cite{Huang2019-learningToPaint} starts with a black canvas.}
    \label{fig:constraint_results}
\end{figure}

With existing SBR RL, it is often not clear what the models are painting until the last set of brush strokes are performed as seen in Figure \ref{fig:constraint_results}.  In this way, the SBR RL models are similar to printers and deterministic SBR models.  In generative art such as GANs \cite{Elgammal2017-CAN}, only the final output image is used for evaluating models.  In machine painting, however, the whole performance needs to be evaluated in addition to the final canvas.  In this paper, we view painting as a plan that is a sequence of actions (or brush strokes) towards an end output. In this context, the proposed idea is to improve the quality of a painting plan according to human-like painting techniques where a subject of the painting becomes incrementally more recognizable as brush strokes are added.

One approach for guiding a RL painting model to paint more like a human would be to collect brush stroke level data from real human painters.  This method would require collecting a large number of samples, each of which would take a long time to produce.  The samples would need to be diverse so that the model generalizes on multiple different types of paintings such as portraits or landscapes.  All-in-all it would be an expensive, time-consuming process.  

Our research goal in this paper is to make the brush stroke planning of a reinforcement learning painter model more human-like.  We focus on the common painting technique of ``blocking-in" where the subject of a painting is recognizable early in the painting process.  We introduce Content Masked Loss which gives more reward to painting regions of the canvas that are important for recognizing the subject matter of the painting.  An object recognition model is used to extract which regions of the image are important for recognition.  Amazon Mechanical Turk workers found that the subject matter of the paintings produced from our models that used Content Masked Loss was apparent far earlier in the brush stroke planning than baseline models using adversarial or $L_2$ loss.

\section{Related Work}

Stroke-based Rendering (SBR) is the process of layering individual elements, such as shapes or brush strokes, onto a digital canvas in order to reproduce a given image.  The goal of SBR may be to perfectly recreate an image with a given set of tools, or it may be to add creative abstractions while rendering.

Recent methods attempt to tackle SBR using deep learning \cite{Ganin2018-spiral,Huang2019-learningToPaint,Nakano2019-neuralPainters,Xie2012-artistAgent}, though the field has existed long before these tools became popular \cite{Hertzmann2003-strokeBasedRenderingSurvey}.  ``Artist Agent" \cite{Xie2012-artistAgent} was an approach to modelling brush strokes on paper using a reinforcement learning agent.  The agent designed brush strokes and received feedback as to whether the strokes were believably made by a human or not. The individual brush strokes produced by Artist Agent were believably human but in order to render a whole painting, the stroke planning was manually produced by humans.

Other SBR work attempts to create an agent that can produce whole paintings rather than single brush strokes. SPIRAL \cite{Ganin2018-spiral} used reinforcement learning to generate brush strokes that would render into a painting that minimized the difference between the canvas and the given image. Work that builds off of SPIRAL, such as \cite{Nakano2019-neuralPainters,Zheng2019-strokenet,Huang2019-learningToPaint}, often use the world model concept \cite{Ha2018-worldModel} in order to speed up training.  In the case of painting, the world model is called a Neural Renderer.  The Neural Renderer is implemented with a differentiable neural network so that the loss can be back-propagated through it.  \cite{Huang2019-learningToPaint} achieves the best results at decomposing an image into a series of strokes. The objective of their reinforcement learning algorithm is to minimize the difference between the generated painting and the given painting.  The authors find that using WGAN loss known as \textit{Wasserstein-l} distance, or \textit{Earth-Mover} distance, generates better results than the $L_2$ distance.  The results from this paper are very successful, for with enough strokes, the model is able to almost completely recreate the input image, but the planning process shows no indication of what the subject matter of the painting is.

Existing datasets that have paired images with strokes lack quality for painting portraits. For instance, the \texttt{QuickDraw} \cite{Ha2017-sketchrnn-quickdraw} dataset, which contains cartoon sketches that users create while trying to render a given image or noun, contain both the brush strokes and image data, however, these sketches are too simple for learning real painting. SketchRNN \cite{Ha2017-sketchrnn-quickdraw} does use paired data, but it can only draw objects that have been labelled in its training data.  This dataset encodes the planning method for how humans sketch, but we need the method of how humans paint.

Reinforcement Learning models such as \cite{Huang2019-learningToPaint} and SPIRAL \cite{Ganin2018-spiral} learn through trial and error how to paint images.  The reward functions are designed to make the model learn which strokes will decrease the loss between the painting and the target image.  Because these models learn without human data guidance, they do not encode how a human painter would paint.  In this paper, we introduce Content Masked Loss which increases the human-like brush stroke planning abilities of the RL model without the use of human painter data.

The brush strokes used in the planning of \cite{Huang2019-learningToPaint} are not capable of being reproduced by a standard paint brush, as seen in Figure \ref{fig:constraint_results}.  Because these brush strokes are not reproducible by a real paint brush, we cannot evaluate if the brush stroke planning is human-like or not.  Constraining the model to use only realistic brush strokes will make it possible to evaluate whether the planning if human-like or not.  In this paper, we constrain our Reinforcement Learning method of Stroke-Based Rendering to produce brush stroke instructions that are realistic and can easily be fed to a painting agent to be painted accurately.  An example of feeding these paint brush instructions can be seen in Figure \ref{fig:robot_painting_example}.

\section{Approach}
A robot painting problem is informally defined as follows. Given an input image and a blank canvas, the goal is to generate a sequence of brush strokes that will turn the canvas into a recreation of the input image, following a human-like painting style. This section describes an RL formulation of the task, focusing on how the proposed reward functions induce human-like plans. 

\subsection{Reinforcement Learning Painter Model}

The proposed work builds on the deep reinforcement learning framework from \cite{Huang2019-learningToPaint}. The original framework is based on several unrealistic assumptions; We modify the framework to address realistic constraints.
First, the brush strokes are allowed to be unrealistically large, e.g., more than half the width of the canvas as seen in Figure \ref{fig:constraint_results}. We constrain what is possible for the brush strokes based on what our robot painting agent is able to do with a single paint brush.  In this way, we can easily feed the brush stroke instructions to a robot such as the one used in our experiments shown in Figure \ref{fig:robot_painting_example}. 
Specifically, we constrained them to be fixed width and only 5\% of the width of the canvas which is consistent with the robot's brush width.  
Second, it was assumed that the blank starting canvas is black which is not consistent with the white paper or canvas that a painter would generally use; we changed this assumption to be white in our model. 
Third, a paint brush can only hold so much paint, so we add the maximum length of one half of the canvas width to the brush strokes based on the robots attempts at painting on a 20cm piece of paper.  
Additionally, since the acrylic paint that the robot utilizes is opaque, we add this as a constraint for the purpose of our study.

We formulate painting as a reinforcement learning task, following the problem formulation in~\cite{Huang2019-learningToPaint}. Here, states represent the canvas from blank to completion. The action space describes the brush strokes in the form of Bézier curves with values for three position coordinates, thickness of the brush stroke at both ends, opacity of the paint at both ends, and color.  The Neural Renderer is the transition function which paints a brush stroke in this action space onto a given canvas.

As the action space is continuous, our model adopts the Deep Deterministic Policy Gradient (DDPG) \cite{lillicrap2016-ddpg}.  The reward function is calculated by computing the difference between the losses at time $t$ and $t+1$ as seen in Equation \ref{eq:reward}.  Huang \textit{et al.} uses an adversarial model to compute the loss function (Equation \ref{eq:gan_loss}).  The model based DDPG critic predicts an expected reward function:
\begin{align*}
    V(c_t) = r(c_t, a_t) + \gamma V(c_{t+1})
\end{align*}
where $r(c_t, a_t)$ is the reward of performing action $a_t$ on canvas $c_t$, and $\gamma$ is a penalty parameter which controls how influencial future value is. $c_{t+1}$ can be renderered using the Neural Renderer to apply the brush stroke $a_t$ to the canvas $c_{t+1}$, seen below as $trans(c_t, a_t)$. The actor in the DDPG, $\pi(c_t)$, is trained to maximize the expected reward:
\begin{align*}
    r(c_t, \pi(c_t)) + \gamma V(trans(c_t, \pi(c_t)))
\end{align*}

\begin{figure}
    \begin{align}
        Reward &= mean \left[ L(c_t, y) - L(c_{t+1}, y) \right] \label{eq:reward} 
    \end{align}
    \centering \textbf{Baseline }
    \begin{align}
        L_{GAN}(c, y) &= 1 - Discriminator(c, y) \label{eq:gan_loss} \\
        L_{L2}(c, y) &= (c - y)^2 \label{eq:l2_loss}
    \end{align}
    \centering \textbf{Ours }
    \begin{align}
        L_{CL}(c, y) &= (VGG_l(c) - VGG_l(y))^2 \label{eq:cl_loss} \\
        L_{CM+L2}(c, y) &= (c - y)^2 * norm(VGG_l(y)) \label{eq:cm_loss} \\
        L_{L1^*}(c, y) &= min( | c - y |, \lambda)  \label{eq:l1_loss} \\
        L_{CM+L1^*}(c, y) &=  min( | c - y |, \lambda) * norm(VGG_l(y)) \label{eq:cm_l1_loss}
    \end{align}
    \caption{The reward function is the per-pixel average of the loss between the canvas at time $t$ and the canvas with an additional stroke at time $t+1$.  The various loss functions used in this paper follow, where $c$ is the canvas and $y$ is the target image.  The $Discriminator$ is a model used to differentiate paintings of an image and the actual image, and $norm()$ is a 0-1 normalizing function.}
    \label{fig:reward_equations}
\end{figure}

\subsection{Reward Functions}

The \cite{Huang2019-learningToPaint} model used the adversarial loss as seen in Equation \ref{eq:gan_loss} and is used in the baseline model in this paper. The $Discriminator$ model outputs 1 if the canvas and target image are the same and 0 if they are completely different. The $Discriminator$ is only trained to determine visual differences between the paintings and the target image so it does not factor in content and abstract detail from the images.

Painters take an abstract approach when painting and consider the subject of the target image.  Human painters use techniques such as ``blocking-in" \cite{Scott2017-blockingIn} which ensure that the subject of the image is recognizable early in the painting process.  We took two different approaches to increasing the recognizability of the painting strokes: Content Loss Reward and Content Masked Reward.

\subsubsection{Content Loss Reward}

In Content Loss (Equation \ref{eq:cl_loss}), we used the $L_2$ loss on the features extracted from the canvas and target images instead of the images themselves.    We used the VGG-16 \cite{Simonyan2015-vgg16} model to extract the features.  The VGG-16 parameters were fixed.  The intuition behind Content Loss is that if a painting and the target image have similar features extracted from an object detection model, then they should recognizably be paintings of the same object.

The VGG-16 \cite{Simonyan2015-vgg16} object detection model was trained on photographs, so it is not expected to generalize to paintings.  We attempt to use Content Loss but recognize that a feature extractor that is robust to both photographs and paintings may be necessary to make this loss function work properly.

\subsubsection{Content Masked Loss Reward}

\begin{figure}[b]
    \centering
    \includegraphics[width=\columnwidth]{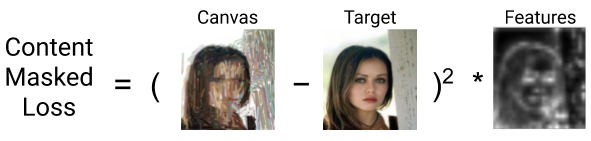}
    \caption{Content Masking added to the $L_2$ loss is defined as the difference between a canvas and the target image weighted by the features extracted from the target image.}
    \label{fig:content_masked_loss}
\end{figure}

Content Masked Loss was designed to increase the weight of the per-pixel loss in ``important" regions of the canvas.  The ``importance" of a region is determined by extracting features using the VGG-16 \cite{Simonyan2015-vgg16} model.  As seen in Figure \ref{fig:content_masked_loss}, the Content Masking can be added to the  $L_2$ distance to weight the difference between a canvas and the target image.  The features are computed by feeding the target image into the VGG-16 model, taking the output of an intermediate layer, averaging the image across the channels, and then 0-1 normalizing it.  We experimented using the output from different layers in the VGG-16 model in Figures \ref{fig:vgg_features} and \ref{fig:content_masked_layers}.

\subsubsection{L1 with Loss Clipping}

Certain biases of the painting model may cover up the effects of Content Masking.  All of the loss functions except Content Loss rely on measures of distance between pixels in the canvas and the target image.  Since the images are in RGB format, the largest distance would be between a white and black pixel.  The canvas is initially white, so the model will have a bias towards painting darker colors, since it can get more reward with them.  We reduced this bias by experimenting with $L_1$ instead of $L_2$ losses.  We also set a constant maximum value for the per-pixel loss. Loss Clipping used in a loss function can be seen in Figure \ref{fig:reward_equations} with $L_1$ ($L1*$) and in conjunction with Content Masked Loss ($CM + L1*$).

\begin{figure}
    \centering
    \includegraphics[width=\columnwidth]{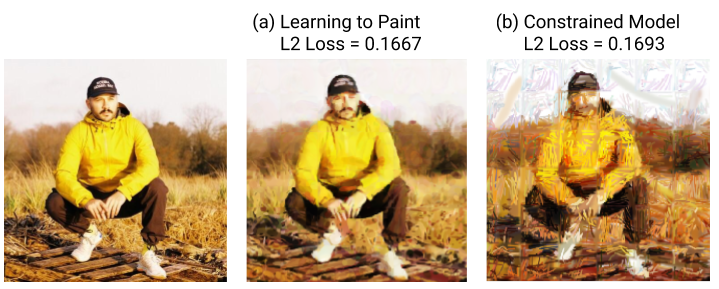}
    \caption{On the left is the target image. The paintings were created by the model generating 125 strokes looking at the whole canvas and then 125 strokes in each of 25 evenly distributed subsections of the canvas. (a) was created using the \cite{Huang2019-learningToPaint} model as is while (b) used constraints on the brush's abilities.}
    \label{fig:fidelity}
\end{figure}

\section{Results}

The effects of the brush stroke constraints (fixed brush width, maximum brush stroke length, fully opaque strokes, and starting with white canvas) can be seen in Figure \ref{fig:constraint_results}.  Our baseline model is able to reproduce the target image but not with as high fidelity as the \cite{Huang2019-learningToPaint} model.  The strokes are now fixed width and smaller than what the \cite{Huang2019-learningToPaint} model allows, so it takes more strokes to reproduce the target image in our baseline model.  The fixed width strokes are too big to make out details such as mouth and eyes.  In order to paint with a smaller brush size, the canvas is divided up into subsections, then the model is run over each subsection.  So if we divide an image into quadrants and run our baseline model over each, this is equivalent to painting with a brush half the width of fixed width constraint.

The last canvases in Figure \ref{fig:constraint_results} show a significant difference in fidelity between our baseline model and the \cite{Huang2019-learningToPaint} model.  One of the most impressive aspects of the \cite{Huang2019-learningToPaint} model was that it could nearly replicate an image given enough strokes.  We tested whether our baseline model could also achieve this accuracy. In Figure \ref{fig:fidelity}, the models generated 125 strokes using the entire image as input, then 125 strokes for each non-overlapping 1/25th of the image.  Our baseline model produces a painting with a similar loss to the \cite{Huang2019-learningToPaint} model's painting.

\subsection{Content Loss Reward}

\begin{figure}[h]
    \centering
    \includegraphics[width=\columnwidth]{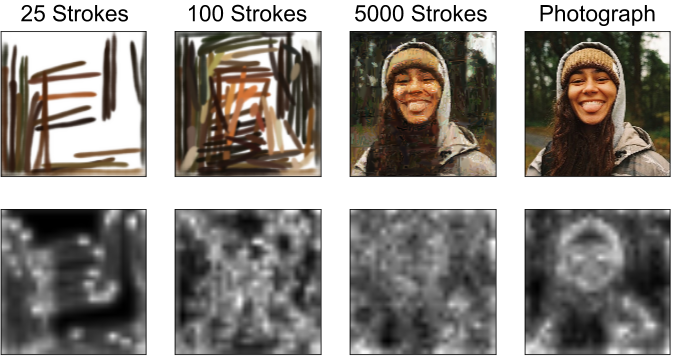}
    \caption{The top left three images are output from the model painting the target image in the top right.  The second row are the features extracted from their corresponding image above them using the first 17 layers of the VGG-16 model.}
    \label{fig:content_loss_problem}
\end{figure}

We tried to use this content loss (Equation \ref{eq:cl_loss}) reward function, but the model would only go to a local minimum and produce blank canvases no matter what parameters we used. The main reason why the Content Loss Reward function was unsuccessful was the feature extraction process does not translate well to the paintings.  The VGG-16 model \cite{Simonyan2015-vgg16} was trained on photographs.  The canvases (especially with only a few strokes) are so vastly different from photographs that the extracted features appear random.  As seen in Figure \ref{fig:content_loss_problem}, even when the painting is extremely similar to the target image, the features extracted are very different from those extracted from the photograph.  The features extracted from the photographic target image make semantic, abstract sense while the features from the paintings are unintelligible.  In future work we will attempt to use feature extraction models that are robust to texture variance, such as \cite{geirhos2018-textureBias} who minimized the texture bias in an ImageNet \cite{deng2009-imagenet} classifier model.

\begin{figure}
    \centering
    \includegraphics[width=\columnwidth]{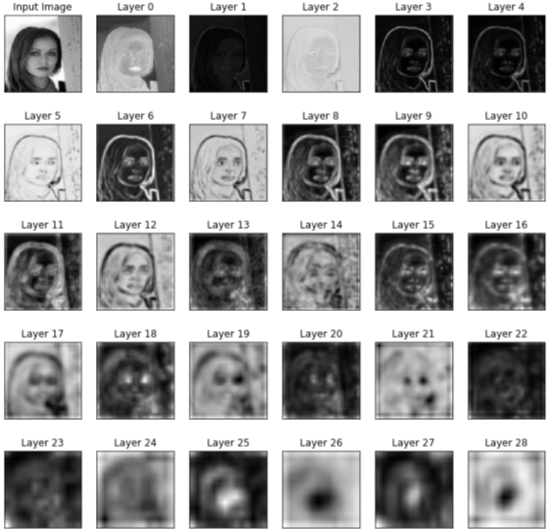}
    \caption{Features extracted from a given image at various layers within the VGG-16 \cite{Simonyan2015-vgg16} model.}
    \label{fig:vgg_features}
\end{figure}

\begin{figure}
    \centering
    \includegraphics[width=\columnwidth]{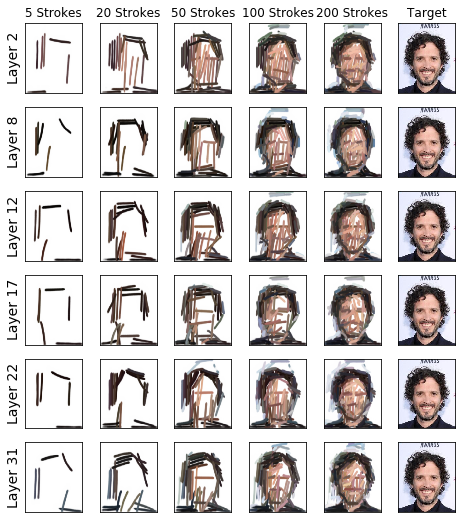}
    \caption{Models were trained using Content Masked Reward where the features were generated using various layers in the VGG-16 model.}
    \label{fig:content_masked_layers}
\end{figure}

\subsection{Content Masked Reward}

The VGG-16 model has has 13 convolutional layers each with an activation function and 5 pooling layers.  These 31 layers can be used in the feature extraction portion of the architecture.  To compute the feature mask, the output of a layer is averaged across channels and 0-1 normalized.  The output of each of the first 29 layers of the pretrained VGG-16 model are shown in Figure \ref{fig:vgg_features}.  The initial layers extract features such as edges and texture, while the later layers extract higher level information such as eyes and mouths.

We trained the model using the Content Masked Loss (Equation \ref{eq:cm_loss}) in the reward function using various layers from the VGG-16 model.  After the models were trained, the painting process for each model was tested in Figure \ref{fig:content_masked_layers}.  The lower level layers produced models that focused more broadly on the subjects face while higher layers were focused on abstract features such as eyes.  All models performed very similar in their final results, however, the intermediate strokes varied.  The models that used latter layers performed more human-like since they focused on the high level features of the image.

Layer 17 was chosen for future experiments since it was a good balance between using high and low level features.

\subsection{Final Painting Quality}

We wanted to test whether there was a difference in quality of paintings amongst the models trained with different reward functions.  We used both the Fréchet Inception Distance (FID) \cite{Heusel2017-fid} and Inception Score \cite{Salimans2016-inception-score}.  
%These measures have known issues \cite{Barratt2018-noteInceptionScore} such as using the metric on data that is not from the ImageNet \cite{deng2009-imagenet} dataset, however, most of the problems occur when comparing the Inception Scores of vastly different models.  
These measures have known issues~\cite{Barratt2018-noteInceptionScore}, e.g., Inception Score, which uses a network pre-trained on ImageNet~\cite{deng2009-imagenet}, can be misleading when used on the models trained on non-ImageNet datasets.  
Since we are merely comparing reward functions and the models have otherwise been trained in the exact same fashion, we can use these scores for comparison.

We computed the FID and Inception Scores for 2000 held out images from the CelebA dataset.  The FID and Inception Scores are shown in Table \ref{tab:fid_and_inception_score}.  They were computed using 5 fold cross validation on 2000 held out images from the CelebA dataset.  We note that both FID and IS scores from our experiments seem far outside the normal range of those scores found in GANs literature \cite{Ostrovski2018-gansAndScores}; the reason for this is that the generated output in our case are paintings that are visually very different from photorealistic input images as in other works. There are significant differences in the FID and IS values between certain models, however, there is no model that is significantly superior to the others in terms of both FID and IS.  The lack of a significantly superior model indicates that the final painting quality of each model is roughly the same, and therefore, the intervention of loss functions does not significantly affect the final painting quality.

\begin{table}[]
    \begin{tabular}{ccc}
    \textbf{Loss Function} & \textbf{FID} $\downarrow$               & \textbf{IS} $\uparrow$  \\ \hline
    GAN           & 240.05 $\pm$ 0.43 & 3.05 $\pm$ .08 \\
    L2            & 241.02 $\pm$ 0.24 & 3.05 $\pm$ .14 \\
    CM + L2            & 242.67 $\pm$ 0.24 & 3.01 $\pm$ .04 \\
    L1*           & 241.58 $\pm$ 0.14 & 3.45 $\pm$ .09 \\
    CM + L1*      & 243.13 $\pm$ 0.20 & 3.47 $\pm$ .05 \\ \hline%
    \end{tabular}%
    \caption{A small Fréchet Inception Distance (FID shown in column 2) indicates that the generated paintings are visually similar to the photographs. The larger the Inception Score (IS shown in column 3), the more similar the distributions of the paintings and target images are.}%
    \label{tab:fid_and_inception_score}
\end{table}

\subsection{Human-Like Planning}

\begin{figure}
    \centering
    \includegraphics[width=\columnwidth]{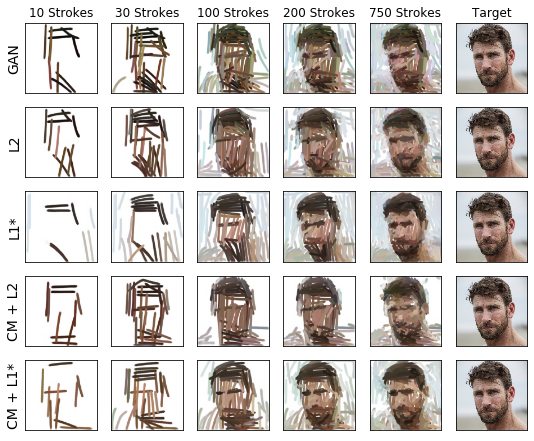}
    \caption{The painting process for models trained using different loss functions in their reward functions.}
    \label{fig:stroke_by_stroke}
\end{figure}

As a human painter paints, the subject of the painting should be apparent per the common ``blocking-in" \cite{Scott2017-blockingIn} technique. As seen in Figure \ref{fig:stroke_by_stroke}, the loss function used to train the painter model affects the intermediate outputs of the model.  For instance, the GAN, $L1^*$, and $L2$ Loss models show that the painting model is not influenced by the content in the target image.  The CM + L2 and CM + L1* models add incentive to paint regions of the image that are important for understanding the content of the image, and so observers can recognize what the model is painting early in the stroke sequence.

We hypothesize that our model, aiming at a human-like planning, produces stroke sequences that can be recognized from early strokes. To verify this hypothesis, we conduct two experiments: first, we use a facial detection AI model on the canvas as the model painted; and second, we ran a study on Amazon Mechanical Turk for human evaluation. 

We note on the difference between AI and human perception.  In general, humans tend to focus on shape-based features in recognizing objects, whereas Convolutional Neural Networks including FaceNet~\cite{Schroff2015-facenet} used in our experiment exhibit strong texture biases~\cite{geirhos2018-textureBias}. Due to the nature of binary classification used in the experiment, the two experiments are not designed in the identical format, i.e., binary classification vs. ranking; however, the results agree that at early stage of under 200 strokes the proposed approach generates paintings where the subject is more easily recognizable when compared to the baseline approaches. 

We tested the following five models, each varying with only the loss function used to compute the reward:
\begin{itemize}
    \item GAN - Adversarial Loss. Equivalent to \cite{Huang2019-learningToPaint} but with brush stroke constraints (Equation \ref{eq:gan_loss}).
    \item L2 - $L_2$ Loss (Equation \ref{eq:l2_loss}).
    \item L1* - $L_1$ with Clipping Loss (Equation \ref{eq:l1_loss}).
    \item CM + L2 - Content Masked $L_2$ Loss (Equation \ref{eq:cm_loss}).
    \item CM + L1* - Content Masked, $L_1$ with Clipping Loss (Equation \ref{eq:cm_l1_loss}).
\end{itemize}

\begin{figure}
    \centering
    \includegraphics[width=\columnwidth]{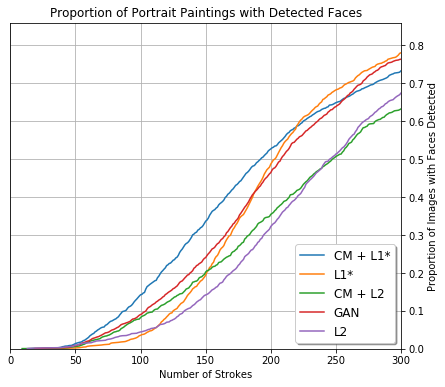}%
    \caption{The proportion of paintings produced by each model that had faces detected by a model \cite{Schroff2015-facenet} within the specified number of brush strokes.}%
    \label{fig:ai_face_detection_results}%
\end{figure}

\subsubsection{AI evaluation.}
In our AI test, each model painted 2000 portraits from a subset of data held out from training.  After each brush stroke, the painting was tested by the FaceNet facial detection model.  The proportion of paintings with detected faces within a certain number of strokes is plotted in Figure \ref{fig:ai_face_detection_results}. Adding content masking generally improved facial detection in early strokes (50-200), although Content Masking's effect on L2 was not strong, likely due to the color bias concealing it. The proposed CM + L1* model produced the most face detected paintings.

\subsubsection{Human evaluation.}
\begin{figure}[t!]
    \centering
    \includegraphics[width=7.2cm]{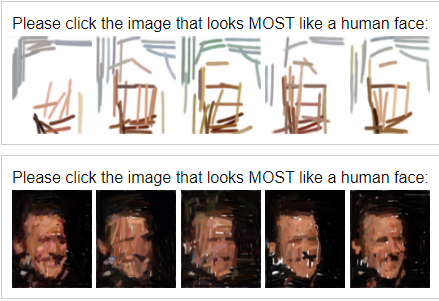}
    \caption{Example of questions in a HIT in our study carried out on Amazon MTurk. Workers viewed paintings by each of the models trained with different loss functions and selected the painting that looked most like a face.}
    \label{fig:amt_example}
\end{figure}

Each model painted the images stroke by stroke.  We showed the workers the canvases produced from each of the five models at various points in the painting process: 10, 30, 100, 200, and 750 strokes. An example of what the Mechanical Turk workers saw in the study is shown in Figure \ref{fig:amt_example}. The order of algorithms was randomized, as was the order of the number of strokes in each image. 332 unique MTurk workers evaluated images.  400 images were used for the paintings, and each question (sample questions are shown in Figure \ref{fig:amt_example}) was evaluated by 4 different workers.

\begin{figure}[t!]
    \centering
    \includegraphics[width=\columnwidth]{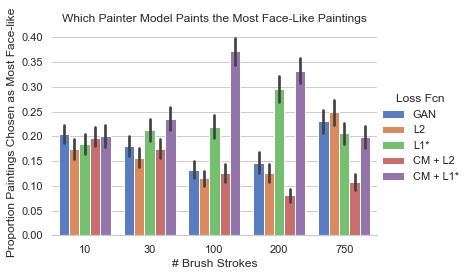}
    \caption{Results from the MTurk study.  Workers chose which painting (performed using 10, 30, 100, 200, or 750 brush strokes) looks most like a face given five paintings from each model.}
    \label{fig:amt_results}
\end{figure}

The results of the study are shown in Figure \ref{fig:amt_results}. Adding Content Masking to L1* loss improved the face-like qualities of the paintings that contained between 30 and 200 brush strokes.  Adding Content Masking to L2 did not significantly improve the face-like qualities of the paintings, and we suspect that this is because of the dark color bias since L1, which is less affected by a color bias, with Content Masking was successful.  Overall, Content Masking with L1 loss proves to create the painter model that produces the most face-like paintings through the painting process.

\section{Conclusion and Future Work}

This paper presents a reinforcement learning approach towards a human-like brush stroke planning, focusing on the procedure of the painting in addition to the final quality. 
Experimental results from both AI and human evaluations show that adding Content Masking to the loss functions greatly improves the facial detection abilities of the model's stroke sequence, without sacrificing the quality of the final painting results. 
Content Masked Loss provides a data-efficient way to increase the human-like abilities of the model's brush stroke sequence without needing additional, expensive human labeling.

We found that the features generated by the VGG-16 object detection model correlate with facial features as seen in Figure \ref{fig:vgg_features} which allowed us to use these features to weight the loss in Content Masking.  
%\cite{geirhos2018-textureBias} points out that convolutional neural networks have a texture bias.  
%Models such as VGG-16 learn to predict objects based on texture rather than the shapes within the image.  
In future work, in order to address the texture bias issue prevalent in CNN-based perception models, we plan to explore using more robust object classifiers as tools for producing the weight matrix in Content Masking that may correlate better with human evaluation.

\section*{Acknowledgements}

We thank Zhewei Huang, Wen Heng, and Shuchang Zhou whose code (\url{https://github.com/hzwer/ICCV2019-LearningToPaint}) we modified for the reinforcement learning models in this paper.

Thank you to Nicholas Eisley (\url{http://neisley.squarespace.com/}) for the photographs used in Figures \ref{fig:ours_vs_baseline}, \ref{fig:constraint_results}, \ref{fig:fidelity}, \ref{fig:content_loss_problem}, and \ref{fig:stroke_by_stroke}.%\ref{fig:ltp_strokes}, \ref{fig:constraint_results}, \ref{fig:fidelity}, \ref{fig:content_loss_problem}, and \ref{fig:stroke_by_stroke}.

\bibliography{ref.bib}

\end{document}